%% file: iclr2021_conference.tex
\documentclass{article} 
\usepackage{iclr2021_conference,times}

\input{math_commands.tex}

\usepackage{hyperref}
\usepackage{url}

\usepackage{svg}
\usepackage{graphicx} 
\usepackage{float} 
\usepackage{subfigure} 
\usepackage{tcolorbox}
\usepackage{CJKutf8}
\definecolor{dt}{gray}{0.7}
\usepackage{multicol}
\usepackage{multirow}
\usepackage{booktabs}
\usepackage{multirow}
\usepackage{makecell}
\usepackage{xspace}
\DeclareSymbolFont{extraup}{U}{zavm}{m}{n}
\DeclareMathSymbol{\varheart}{\mathalpha}{extraup}{86}
\DeclareMathSymbol{\vardiamond}{\mathalpha}{extraup}{87}

\newcommand{\modelname}{\textsc{VisCPM}\xspace}
\usepackage{tablefootnote}

\title{\Large Ziya-Visual: Bilingual Large Vision-Language Model via Multi-Task Instruction Tuning}

\author{
 Junyu Lu$^{\varheart}$\footnotemark[1] \qquad
    Dixiang Zhang$^{\varheart\vardiamond}$\footnotemark[1] \qquad
    Xiaojun Wu$^{\varheart}$\footnotemark[1] \qquad
    Xinyu Gao$^{\varheart}$ \qquad
    \\
\hspace{8mm}\textbf{
        Ruyi Gan$^{\varheart\clubsuit}$\footnotemark[2] \qquad
    Jiaxing Zhang$^{\varheart}$ \qquad
    Yan Song$^{\clubsuit}$ \qquad
    Pingjian Zhang$^{\vardiamond}$ \qquad
    }
    \\
    \\
\hspace{2mm}$^\varheart$International Digital Economy Academy \quad
    $^\clubsuit$University of Science and Technology of China \quad \\
    \hspace{-2mm}$^\vardiamond$South China University of Technology \quad 
    \\
    \\
    \hspace{2mm}{\tt\small \{lujunyu, zhangdixiang, wuxiaojun, ganruyi, zhangjiaxing\}@idea.edu.cn } \\
    {\tt\small clksong@gmail.com} \hspace{4mm}
    {\tt\small pjzhang@scut.edu.cn}
}

\begin{document}

\maketitle

{
  \renewcommand{\thefootnote}%
    {\fnsymbol{footnote}}
  \footnotetext[1]{Equal Contribution.}
  \footnotetext[2]{Project Leader.}
}

\input{sec/abs}
\input{sec/introduction}

\input{sec/relate_work}
\input{sec/method}
\input{sec/experiment}

\section{Conclusion and Future Work}

We release the Ziya-Visual series, a comprehensive set of large-scale bilingual vision-language models and instruction-response pair construction pipelines, which are designed to foster development within the multi-modal open-source community. The latest Ziya-Visual-Chat achieves comparable or slightly superior performance to mainstream English-only and bilingual LVLMs across various benchmarks, demonstrating significant multi-modal understanding and generation capabilities in multi-turn conversation, complex reasoning and detail descriptions. Looking forward, we are committed to further enhancing the capabilities of Ziya-Visual in several key aspects:
\begin{itemize}
    \item[$\bullet$]
    Developing the capabilities of Ziya-Visual in multi-modal tasks such as referential dialogues, general scene differentiation, video comprehension.
    \item[$\bullet$]
    Enhancing the fine-grained visual object understanding of Ziya-Visual by extending the model structure and training data, enabling it to handle detail portrait and action recognition within complex images.
    \item[$\bullet$]
    Extending Ziya-Visual's functionality in multi-modal generation, particularly in the integration of diffusion to achieve text-to-image and image-to-image generation.
\end{itemize}

\bibliography{iclr2021_conference}
\bibliographystyle{iclr2021_conference}

\newpage
\appendix
\vspace{2em}
\begin{table}[h]
  \begin{minipage}{0.99\textwidth}
\centering  
\vspace{-4mm}
\scalebox{0.89}{
\begin{tabular}{l p{12.5cm} }
\toprule
 \multicolumn{2}{l}{\bf Visual input example, suitcases:}  \\
\midrule
&  \includegraphics[height=4.5cm]{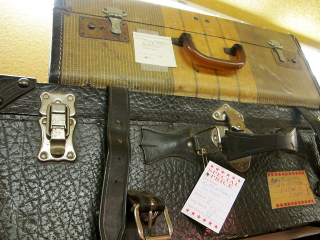} \\

User & What is unusual about this image? \\

\midrule
Ziya-Visual-V1 &  The colors of the two suitcases in the image are different. One is brown, and the other is black.
\\ \midrule
VisualGLM & The two suitcases in the picture are black and brown. They have a white label indicating that they are antiques or vintage items. These boxes may be antique suitcases, travel cases, or part of antique furniture. Their design and decoration may suggest their use in various historical periods.
\\
\bottomrule
\end{tabular}
}
\vspace{1mm}
\caption{Case Study of Ziya-Visual-V1 and VisualGLM.}  
\label{tab:visual_example_bags}  
  \end{minipage}
\end{table}
\subsection{Case Study}
Using Ziya-Visual-V1 (Old name of Ziya-Visual-Base) and VisualGLM as examples and through case analysis, we observed that Ziya-Visual-V1 tends to provide concise and accurate answers, whereas VisualGLM's responses are more abundant and divergent. This observation may explain why Ziya-Visual-V1 outperforms VisualGLM in simple and detailed conversations but slightly falls behind in complex reasoning tasks. Although we haven't conducted detailed ablation experiments, we hypothesize that the primary cause of these differences lies in the variation of image question-answering training data rather than the differences in the employed LLM.

We illustrate our analysis with an example \ref{tab:visual_example_bags} from coco test set: Given the question "What are the colors of the two suitcases in the picture?" The responses from the two models are as follows:

GPT-4 provides a rating of 9 for Ziya-Visual-V1 (Assistant 1) and 7 for VisualGLM (Assistant 2), with the explanation: Assistant 1's response is concise and straightforward, directly answering the question by providing the colors of the two suitcases as black and brown. Assistant 2's response also mentions the colors of the suitcases but includes unnecessary information such as the white label and the possible uses of the suitcases as antiques. While this additional information might be interesting, it does not directly address the question. As a result, Assistant 1's response is considered more useful, relevant, and accurate by GPT-4.

Also, let's now take a look at some examples from Ziya-Visual-V1 to gain an intuitive understanding of its bilingual visual question-answering capabilities.

\begin{figure}[h]
  \centering
  \includegraphics[width=\linewidth]{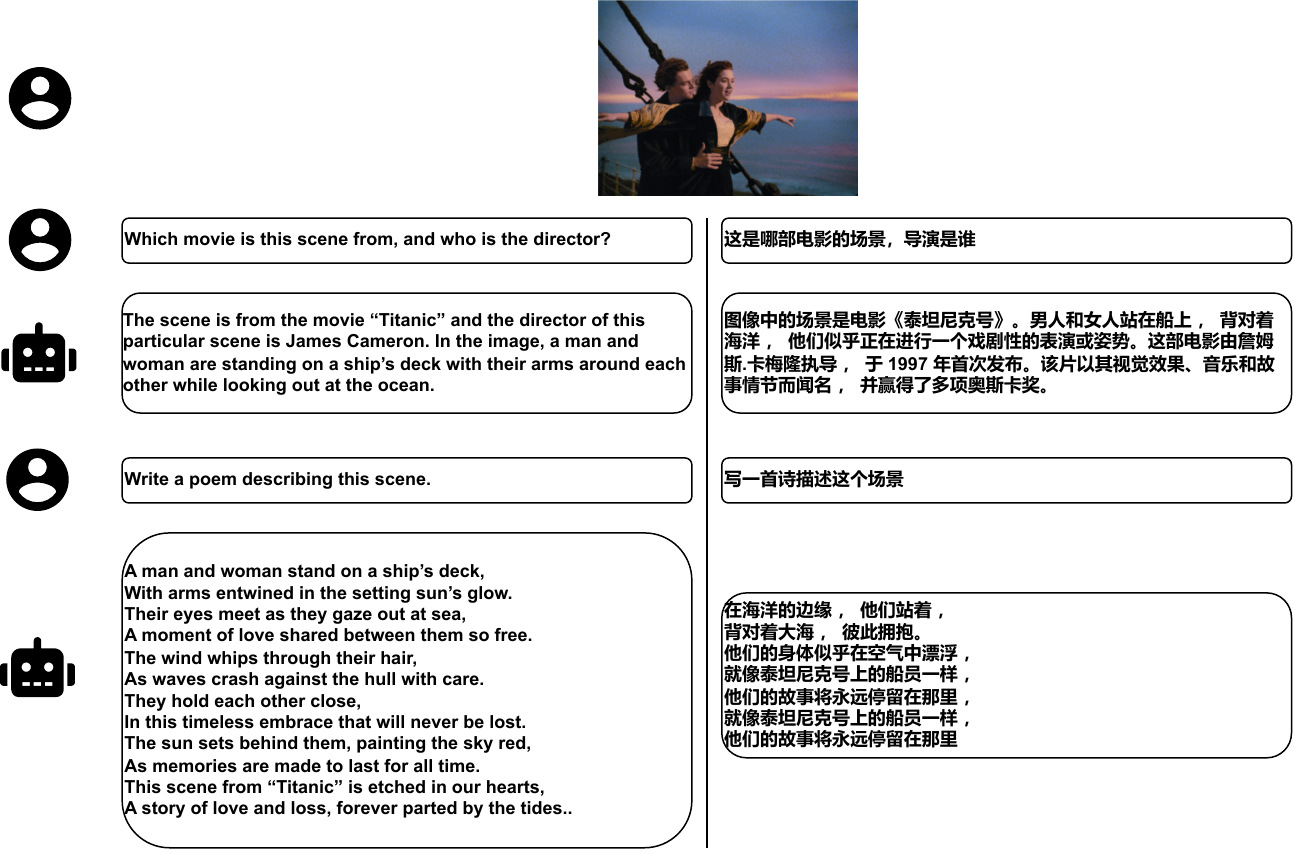}
  \vspace{-1mm}
  \caption{Visual Dialogue Example 1: Titanic Movie Clip. Note that the text on the left and right are generated independently, and the Chinese text on the right is not a translation of the English text on the left.}
  \label{fig:titanic}
\end{figure}

This example \ref{fig:titanic} showcases the model's image recognition, knowledge, and creative abilities. In the first question, the model identifies the scene from the movie "Titanic" in the image and provides information about the film's director, release date, and achievements in awards. In the second question, the model creatively generates a modern love poem based on the user's request.

\begin{figure}[h]
  \centering
  \includegraphics[width=\linewidth]{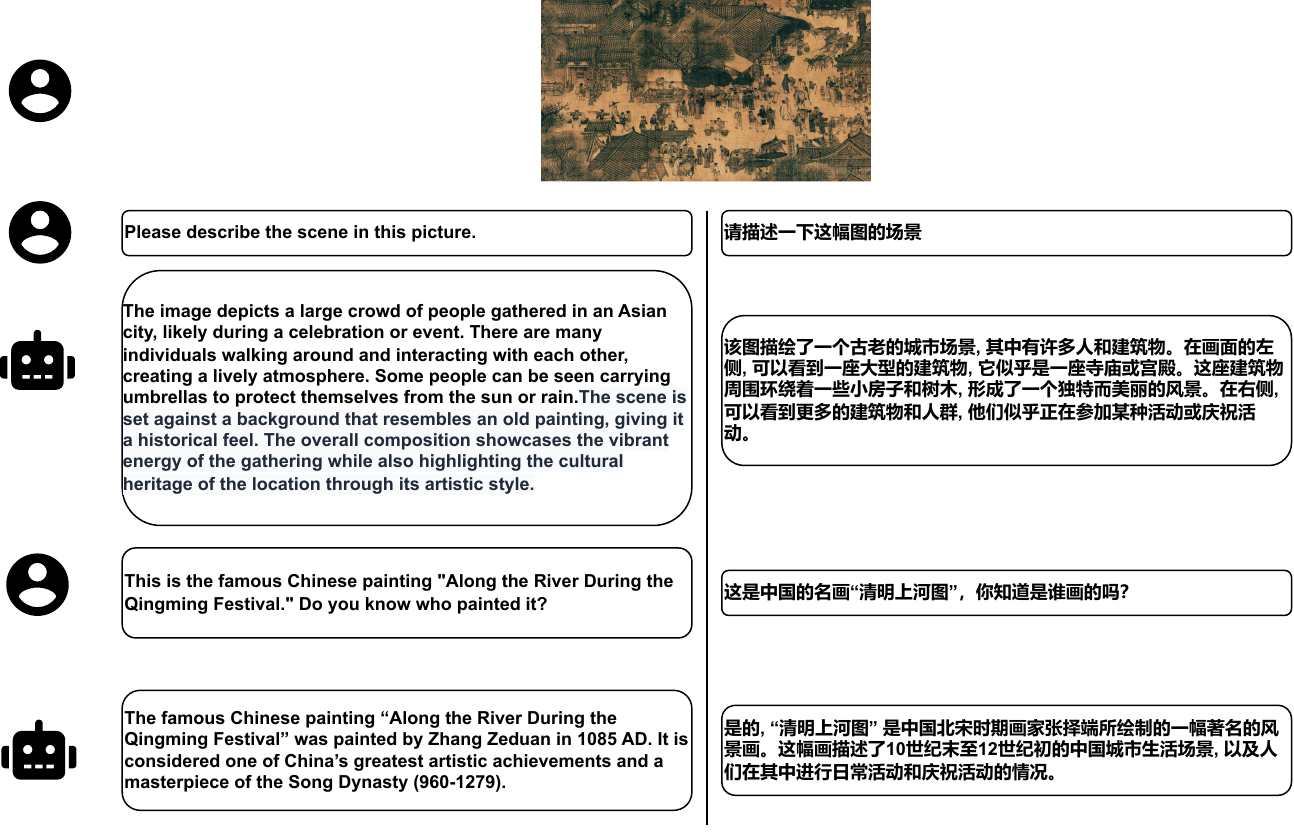}
  \vspace{-1mm}
  \caption{Visual Dialogue Example 2: Ancient Chinese Masterpiece: "Qingming Shanghe Tu" (Along the River During the Qingming Festival).}
  \label{fig:song}
\end{figure}

This example \ref{fig:song} illustrates Ziya-Visual-V1's recognition and understanding of traditional Chinese culture. The model identifies information from a Chinese painting and, upon receiving the clue "Qingming Riverside Picture," provides details about the painter Zhang Zeduan and the historical background of the Northern Song Dynasty.

\begin{figure}[h]
  \centering
  \includegraphics[width=\linewidth]{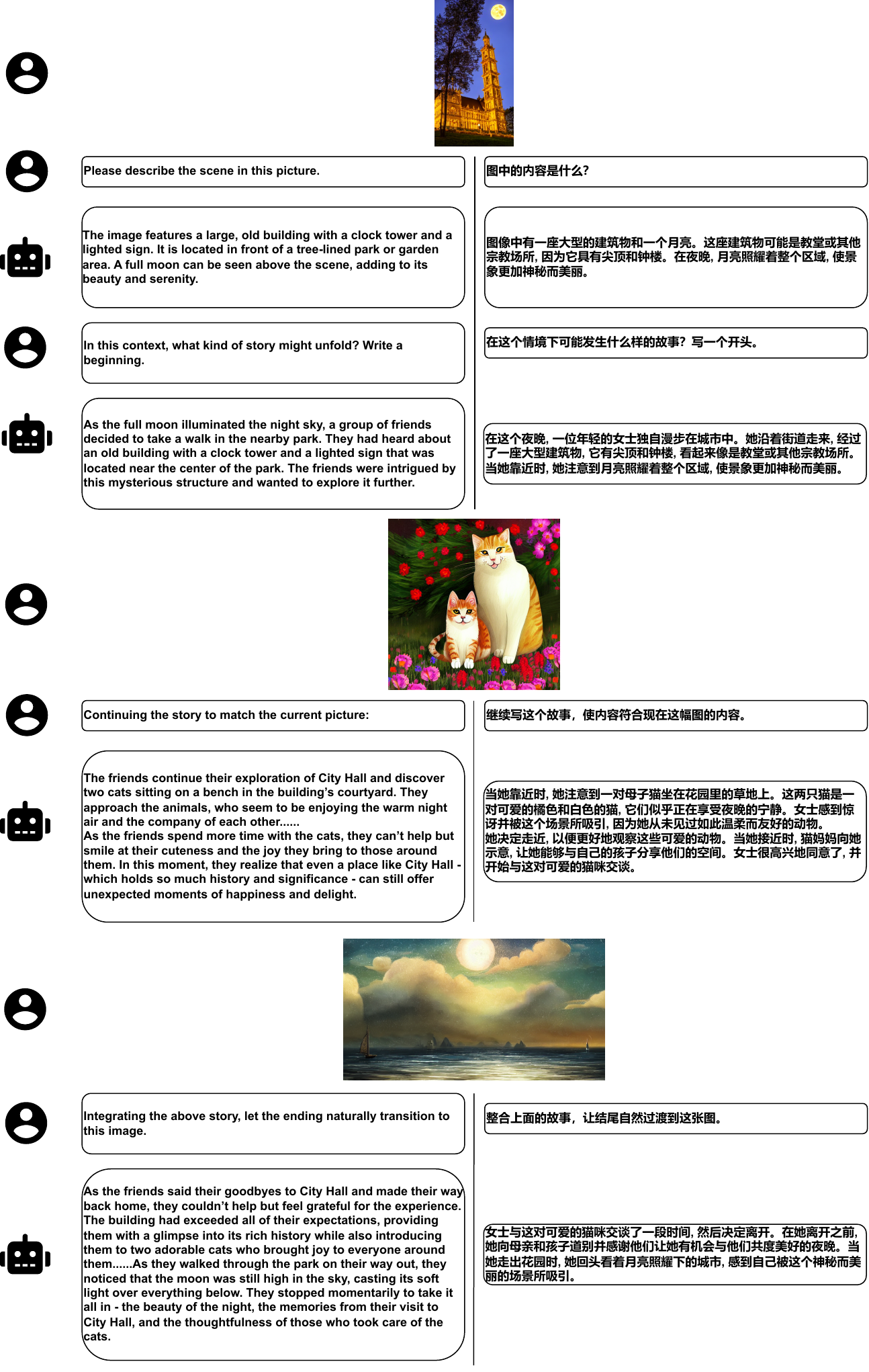}
  \caption{Visual Dialogue Example 3: Multiple Images and Extended Dialogue.}
  \label{fig:multiple}
\end{figure}

Ziya-Visual-V1's competence extends to handling multiple images and conducting multi-turn interactions. In this example \ref{fig:multiple}, Ziya-Visual-V1 demonstrates its powerful capabilities by narrating a story based on three images provided by the user. Thanks to the creativity of the generative model, the storylines in the generated Chinese and English texts are indeed different. The English story depicts a night outing of a group of friends, while the Chinese story is completely distinct. The Chinese story revolves around a lady encountering a mother cat and her kittens amidst the city's night scenery. The story involves interactions and conversations between the lady and the feline family, leading to heartwarming separations.

\begin{figure}[h]
  \centering
  \includegraphics[width=0.5\textwidth]{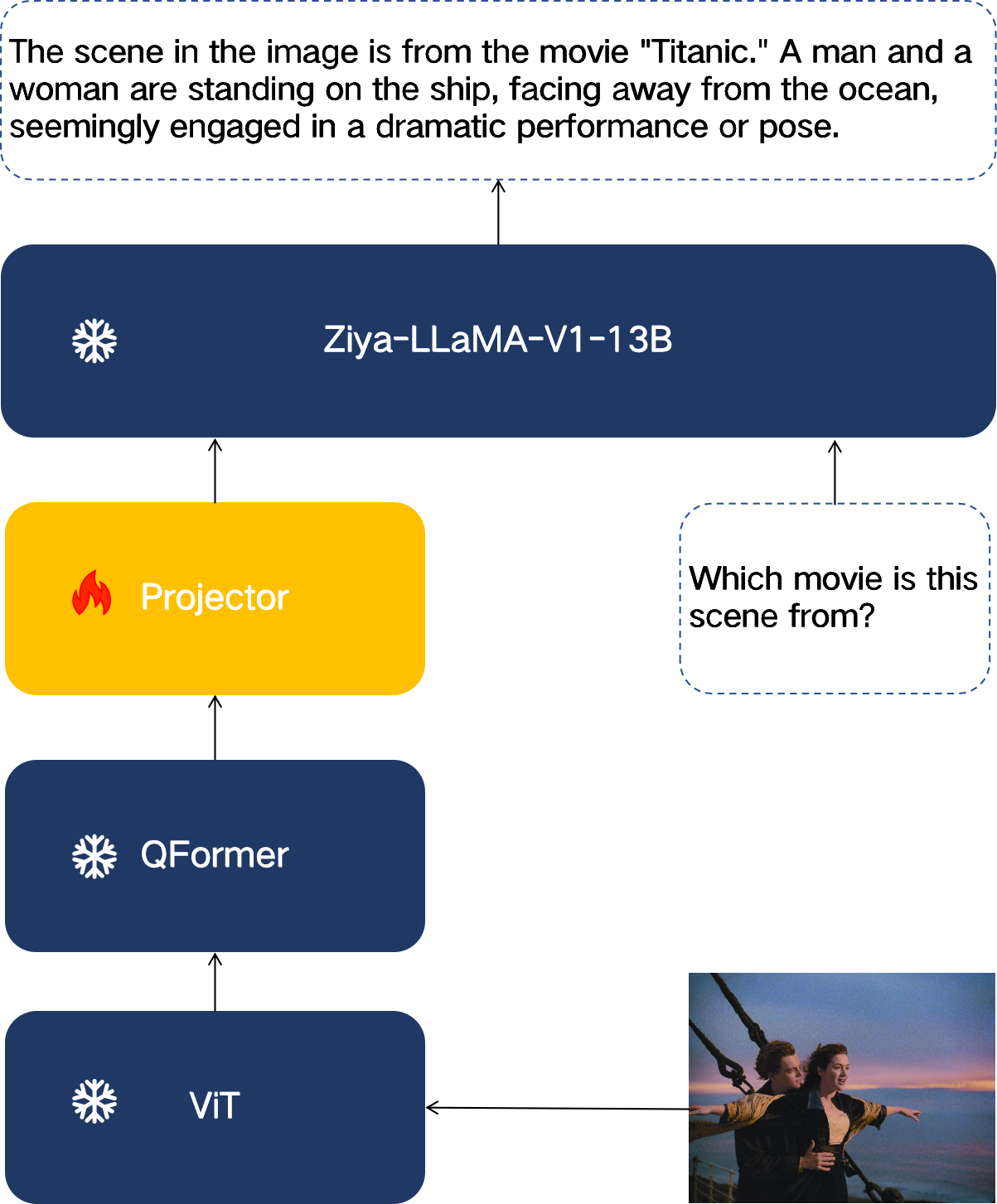}
  \caption{This is an overview of our Ziya-Visual-Base. The basic model structure is a multi-modal Vision Transformer model interacted with Large Language Model. In the process, images are fed into the Vision Transformer (ViT) and Q-Former to extract semantic features (instruction are fed into the Q-former for Ziya-Visual-Chat). These features are then aligned in dimension with the token embeddings of the Large Language Model (LLM) using a projection layer, resulting in Visual Tokens. The Visual Tokens are combined with Instruction and jointly input into the LLM for text generation. This approach enables the model to effectively utilize both visual and textual information in generating coherent and relevant text outputs.}
\end{figure}

\end{document}

%% file: math_commands.tex
\usepackage{amsmath,amsfonts,bm}

\def\eqref#1{equation~\ref{#1}}

\def\1{\bm{1}}

\DeclareMathAlphabet{\mathsfit}{\encodingdefault}{\sfdefault}{m}{sl}
\SetMathAlphabet{\mathsfit}{bold}{\encodingdefault}{\sfdefault}{bx}{n}



%% file: sec/abs.tex
\begin{abstract}
Recent advancements enlarge the capabilities of large language models (LLMs) in zero-shot image-to-text generation and understanding by integrating multi-modal inputs. However, such success is typically limited to English scenarios due to the lack of large-scale and high-quality non-English multi-modal resources, making it extremely difficult to establish competitive counterparts in other languages. In this paper, we introduce the Ziya-Visual series, a set of bilingual large-scale vision-language models (LVLMs) designed to incorporate visual semantics into LLM for multi-modal dialogue. Composed of Ziya-Visual-Base and Ziya-Visual-Chat, our models adopt the Querying Transformer from BLIP-2, further exploring the assistance of optimization schemes such as instruction tuning, multi-stage training and low-rank adaptation module for visual-language alignment. In addition, we stimulate the understanding ability of GPT-4 in multi-modal scenarios, translating our gathered English image-text datasets into Chinese and generating instruction-response through the in-context learning method. The experiment results demonstrate that compared to the existing LVLMs, Ziya-Visual achieves competitive performance across a wide range of English-only tasks including zero-shot image-text retrieval, image captioning, and visual question answering. The evaluation leaderboard accessed by GPT-4 also indicates that our models possess satisfactory image-text understanding and generation capabilities in Chinese multi-modal scenario dialogues. Code, demo and models are available at ~\url{https://huggingface.co/IDEA-CCNL/Ziya-BLIP2-14B-Visual-v1}.

\end{abstract}

%% file: sec/introduction.tex
\section{Introduction}

Large language models (LLMs) such as GPT3~\cite{brown2020gpt3}, LLaMA~\cite{touvron2023llama} and Vicuna~\cite{vicuna2023} have attracted widespread attention due to their powerful text generation and understanding capabilities. These models can demonstrate powerful interactive capabilities by further learning user intentions in carefully designed instruction tuning datasets~\cite{wei2021flan}. Recently, prominent large-scale vision-language models (LVLMs), such as BLIP-2~\cite{li2023blip2}, LLaVA~\cite{liu2023llava}, and mPLUG-Owl~\cite{ye2023mplugowl} have been developed to explore the potential of LLMs in perceiving and understanding visual signals, and they demonstrated impressive capabilities in solving real-world multi-modal dialogues and reasoning.

However, the success of large-scale vision-language models is primarily achieved within the English community, as a large amount of image-text captioning datasets, such as Laion~\cite{schuhmann2022laion}, CC12M~\cite{changpinyo2021cc12m} and SBU~\cite{ordonez2011sbu}, can be used to align textual and visual representations. Furthermore, mainstream studies attempt to uniformly convert supervised multi-modal data from several task categories into instruction tuning formats, such as MULTIINSTRUCT~\cite{xu2022multiinstruct} and MIMIC-IT~\cite{li2023mimic}, which comprehensively bridge the heterogeneity of multi-modal tasks across various scenarios and types. As a result, the data resource gap has severely hindered the development of non-English multi-modal models.

To promote the vigorous development of the multi-modal open-source community, we introduce the open-source Ziya-Visual series. The bilingual Ziya-Visual models include Ziya-Visual-Base and Ziya-Visual-Chat versions, both of which expand the Ziya-LLaMA-13B language model through the Q-Former and visual encoder, endowing it with visual comprehension capabilities. Specifically, Ziya-Visual-Base continues the breakpoints of BLIP-2~\cite{li2023blip2}, aligning vision-language representations under the premise of freezing the LLM and visual encoder. After the pre-training stage and two-stage instruction tuning, it uses a spot of image captioning and instruction-tuning data to stimulate the LLM's understanding and generation capabilities of visual information. Additionally, Ziya-Visual-Chat is an interactive vision-language model based on the pre-trained instructBLIP framework~\cite{instructblip}, which adapts several training strategies, including instruction tuning and low-rank adaptation, to improve bilingual alignment. After multi-stage pre-training and instruction tuning, it can more flexibly extract informative features from images and instructions, and supports complex multi-modal scenarios such as multi-turn dialogues, situational question answering, logical reasoning. To achieve this, we leverage substantial multi-modal English resource as a pivot to present a Bilingual Multi-Modal In-Context (BMMIC) dataset, consisting of over 5 milion image-text pairs, which utilizes GPT-4~\cite{openai2023gpt4} for automated translation and generation of Chinese vision-language question-answer pairs. Specifically, the features of the Ziya-Visual series models include:

\begin{itemize}
    \item[$\bullet$]
    Competitive Performance: It is on par with the state-of-the-art monolingual open-source Large Vision Language Models under the same-level model scale on several evaluation benchmarks, including zero-shot image captioning, visual question answering and visual reasoning.
    \item[$\bullet$]
    Bilingual capability supporting text recognition and visual reasoning: The Ziya-Visual naturally supports multi-modal dialogue in both English and Chinese scenarios. We have explored the impact of various training strategies on model performance, serving as a valuable reference for fellow researchers. 
    \item[$\bullet$]
    Open-Source bilingual multi-modal data: We provide a multi-modal in-context instruction-response dataset encompassing various real-world dialogue scenarios, along with an automation annotation pipeline for translation and generation.
\end{itemize}

%% file: sec/relate_work.tex
\section{Related Work}
\subsection{Multi-modal Large Language Model}
Due to the scaling up of training data and model size, the expansion of large language models to multi-modal learning has attracted widespread attention. Flamingo~\cite{alayrac2022flamingo}, LLaMA-Adapter~\cite{zhang2023llama-adapter, gao2023llama-adapter-v2} and Otter~\cite{li2023otter} integrate learnable cross-attention layers into the pre-trained LLMs to perceive visual knowledge, and train on a large-scaled interleaved image-text dataset. To more efficiently and effectively boost vision-and-language pre-training, BLIP-2~\cite{li2023blip2} uses a Q-Former as perceivers to align queried visual feature with text, which pre-trains the representation learning via multiple vision-language losses. InstructBLIP~\cite{instructblip} further proposes instruction-aware visual feature extraction, which enables flexible and informative feature extraction according to the given instructions. Additionally, some methods transmit the features extracted from visual encoder to the LLM input through linear projection, such as LLaVA~\cite{liu2023llava} and Shikra~\cite{chen2023shikra}, which utilize visual information in a more direct and intuitive way.

\subsection{Multi-modal Instruction Tuning Dataset}
To explore instruction tuning for multi-modal learning, MULTIINSTRUCT~\cite{xu2022multiinstruct} first proposes a multi-modal instruction tuning benchmark dataset, converting 62 different multi-modal tasks into a unified seq-to-seq format. Due to GPT4's demonstrated strong understanding of multi-modal textual representations, LLaVA~\cite{liu2023llava} introduces a data reformation perspective and pipeline to convert image-text pairs into the appropriate instruction-following format by utilizing ChatGPT/GPT-4~\cite{openai2023gpt4}. Furthermore, MIMIC-IT~\cite{li2023mimic} specializes the generation of instruction-response pairs through multi-modal in-context information and diverse visual scenes. M3IT~\cite{li2023m3it} converts classical vision-language tasks into a unified vision-to-text schema via manual instruction writing and dataset pre-processing, including captioning, visual question answering, visual conditioned generation, reasoning and classification.

%% file: sec/method.tex
\section{Methodology}
\subsection{Instruction-Response Construction}

\begin{figure}[h]
  \centering
  \includegraphics[width=\linewidth]{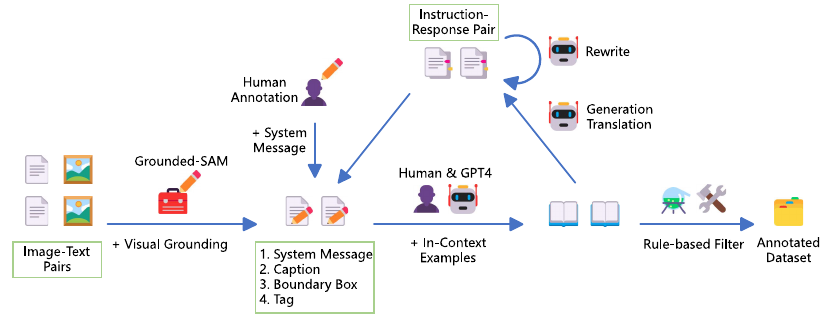}
  \vspace{-1em}
  \caption{The overview of our bilingual instruction-response pairs construction. We use human annotation and a warm-up stage with GPT-4~\cite{openai2023gpt4} to generate system messages and in-context examples. Subsequently, we leverage GPT-4's multi-modal understanding capability to generate instruction-response pairs and the pre-prepared filters to select high-quality data.}
  \label{fig:pipeline}
\end{figure}

In this section, We aim to build a bilingual multi-modal in-context (BMMIC) dataset to support more LVLMs in enhancing the ability to comprehend the real-world dialogue. As illustrated in Figure~\ref{fig:pipeline}, we provide an overview of the BMMIC dataset, including the data preparation and automated instruction-response generation/translation pipeline.

The community has witnessed the emergence of publicly available multi-modal data such as image-text pairs, encompassing referential question answering, contextual dialogues and caption descriptions about images. However, the availability of multi-modal instruction-following data is limited, primarily because the annotation process is time-consuming and annotation targets are less well-defined. Taking inspiration from the impressive performance of GPT-4 in multi-modal association and in-context learning, we select the COCO~\cite{lin2014coco}, Flickr30k~\cite{young2014flickr30k} and AI\_Challenger\_2017~\cite{aichallenger} datasets for multi-modal generation and the LLaVA~\cite{liu2023llava}, M3IT~\cite{li2023m3it} datasets for multi-modal translation through an instruction-following approach. We mainly consider the design of system messages and in-context learning~\cite{liu2023llava, dong2022incontext} when invoking the GPT-4 interface. Further, to mitigate the occurrence of erroneous instruction-response pairs due to cognitive bias in GPT-4, we develop a rule-based detector to filter out data that are out of expectations.

\paragraph{Input Format:}We obtain two types of symbolic representations from annotated multi-modal datasets: (1) \textit{Captions} typically describe the visual scene from various perspectives. (2) \textit{Bounding boxes} localize visual objects within visual scene, with each bounding box encoding the semantics and spatial position of the object. Here, we leverage the Grounded-SAM pipeline, consisting of the Grounding-DINO~\cite{liu2023dino}, SAM~\cite{kirillov2023sam} and RAM~\cite{zhang2023ram} modules, to extract the fine-grained visual feature from images which are further transformed into boundary boxes and visual tags as symbolic representations.

\paragraph{System Messages:}We customize the GPT-4 role for instruction generation and translation based on the composition of instruction types and rewriting requirements, including input format, response types, style constraints and expansion paths. Following LLaVA, we denote the \textit{Conversation}, \textit{Detail description} and \textit{Complex reasoning} to generate instruction data and \textit{Translation} to translate available English instruction-response pairs. We further provide \textit{Deepening} (increase the depth and breadth of the instruction-reponse pairs), \textit{Concretizing} (complex the instruction-reponse pairs by replacing general concepts with more specific concepts, \textit{Increasing Reasoning} (rewrite the responses to explicitly request multiple-step reasoning) and \textit{Adding Constraints} (constraining and specializing the format and content of instruction-response pairs), four rewriting schemes to improve the diversity and richness of instructions and responses.

\paragraph{In-Context Learning:}For the $p$-th image, we take several associated captions $X^C_q$ along with tagged bounding boxes $X^B_q$, and manually annotate 50 instruction-response pairs as query examples, each of which is denoted as $I_q$ and $R_q$. Further, we could define an in-context function $C_\psi: \left(User[X^C_q, X^B_q], Assistant[I_q, R_q]\right) \mapsto p_\theta \left(Assistant[I_k, R_k]\ |\ User[X^C_k, X^B_k]\right)$ to represent the multi-modal in-context generation with current query example. Similarly, we simply switch to the corresponding system instructions for translating the English instruction-response pairs.

\subsection{Model Architecture}

The overall architecture of Ziya-Visual series consists of three components:

\paragraph{Large Language Model:}Ziya-Visual adopt a bilingual large language model as the foundation backbone, which is initialized with pre-trained Ziya-LLaMA-13B~\cite{fengshenbang}. Ziya-LLaMA-13B inherites the breakpoints from the initial LLaMA, adapting the tokenizer, training strategy and model weights for bilingual version.

\paragraph{Visual Encoder:}We adopt Vision Transformer (ViT)~\cite{dosovitskiy2020vit} as the visual encoder to extract a fixed number of output features from the images, initialized with pre-trained weights from ViT/G-14.

\paragraph{Querying Transformer:}To alleviate the issues of information redundancy and inefficiency caused by long image feature sequences, Ziya-Visual uses the Q-Former~\cite{li2023blip2} to bridge the gap between the pre-trained image encoder and LLM. As shown in Figure~\ref{tab:pretraining}, Q-Former consists of two transformer submodules that share the same self-attention layers: (1) an image transformer that interacts with the visual encoder for image features compression. The module uses a group of learnable visual query embeddings as input to the image transformer, which interact with each other through self-attention layers and compress image feature from 256$\times$768 to 64 through cross-attention layers. (2) a text transformer that encodes the text presentations paired with input images. The Q-Former controls the visibility between quries and text through the self-attention mask matrix, enabling the execution of various pre-training tasks to align vision-language features. During instruction-tuning stage, the compressed visual feature sequence is subsequently fed into the large language model.

\subsection{Training}
The training process of the Ziya-Visual models consist of three stages: a first stage of pre-training and two stage of instruction-tuning training. 

\begin{figure}[h]
    \label{tab:pretraining}
  \centering
  \includegraphics[width=\textwidth]{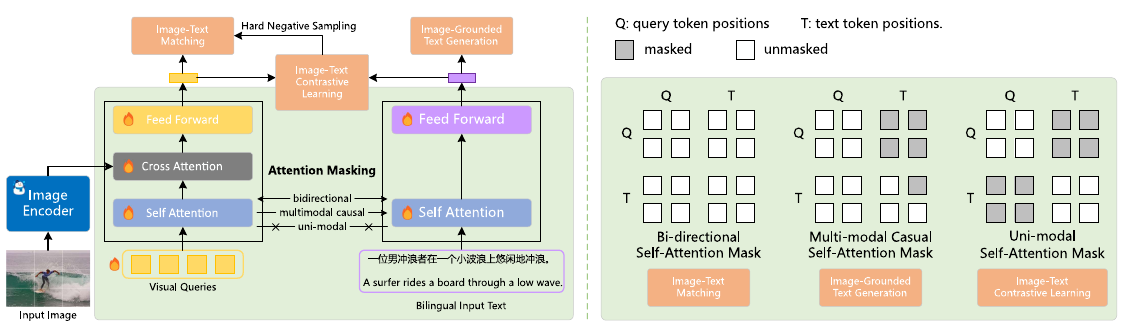}
  \vspace{-1em}
  \caption{\textbf{(Left)} Model architecture of Q-Former and pre-training objectives. We jointly optimize three objectives that incentivize the queries (a set of trainable embeddings) to extract the most pertinent visual representations in alignment with the text. \textbf{(Right)} The self-attention masking strategy for each objective to control query-text interaction. The model architecture and training design are referred from BLIP2~\cite{li2023blip2}.}
\end{figure}

\paragraph{Pre-training:}As illustrated in Figure~\ref{tab:pretraining}, during the pre-training stage, we mainly collect several large-scale, high-noise datasets of web-crawled image-text pairs. We leverage an in-house bilingual CLIP-v2~\cite{radford2021clip} model to clean up and accumulate a portion of high-quality Chinese-English image captioning data based on publicly available open-source datasets, which contributes to improving the overall data quality. As summarized in Table~\ref{tab:pretraining_data}, we prepare a total of 13 million English and 17 million Chinese image-text pairs for pre-training.

\begin{table}[ht]
    \centering
    \caption{Details of Ziya-Visual pre-training data. CC12M~\cite{cc12m}, CC3M~\cite{cc3m}, SBU~\cite{sbu} and COCO Caption are academic caption datasets. Laion$_{High}$ is a Chinese language subset of LAION~\cite{schuhmann2022laion} with high resolution, and Wukong~\cite{gu2022wukong} dataset is a large-scale multi-modality Chinese dataset, provided by Huawei Noah Lab. AI Challenger~\cite{aichallenger} is Chinese competition dataset.}
\vspace{1em}
\scalebox{1}{
    \begin{tabular}{ll ccc}
         \toprule
         \textbf{Language} & \textbf{Dataset} & \textbf{Original} & \textbf{Cleaned} & \textbf{Remaining$\%$} \\
         \midrule
         \multirow{4}{*}{English}
         & CC12M        & 10.5M      & 9M   & 85$\%$ \\
         & CC3M         & 2.8M       & 2.8M   & 100$\%$ \\
         & SBU          & 0.88M       & 0.7M & 80$\%$ \\
         & COCO Caption & 0.6M     & 0.6M & 100$\%$ \\
         \midrule
         \multirow{3}{*}{Chinese} & AI Challenger     & 3M     & 3M & 100$\%$ \\
         & Wukong & 100M & 12M & 12$\%$ \\
         & Laion$_{High}$ & 2.3M & 1.8M & 75$\%$ \\
         \bottomrule
    \end{tabular}}
    \label{tab:pretraining_data}
\end{table}

We connect Q-Former to a frozen image encoder and perform pre-training stage with image-text pairs, aiming to guide the learned queries to extract the most informative visual representations from the text through representation aligning. Following the practices of BLIP2~\cite{li2023blip2} and ALBEF~\cite{li2021albef}, we also jointly optimize three pre-training objectives, Image-Text Contrastive Learning (ITC), Image-grounded Text Generation (ITG) and Image-Text Matching (ITM), which share the same input format and model parameters. Specially, for Ziya-Visual-Chat, We further use the contrastive similarity to sample one in-batch hard negative image-text pair for the ITM task. The model is trained using AdamW~\cite{kingma2014adam} optimizer with learning rate $lr = 1e^{-5}$, $\beta_1 = 0.9$, $\beta_2 = 0.9$ and $eps = 1e^{-8}$. We use a weight decay of $5e^{-2}$ and a gradient clipping of $2.0$. The training process uses a batch size of 2048 for the image-text pairs, and the entire pre-training stage lasts for 90000 steps.

\paragraph{Instruction-Tuning:}In the first stage of instruction-tuning (Multi-task Representation Learning), to ensure diversity of the instruction-tuning data, we collect a wide range of publicly available vision-language fine-tuning datasets and transform them into instruction-tuning format using manually devised rules. As summarized in Table~\ref{tab:multitask_data}, we train Ziya-Visual on 4 tasks simultaneously. For image captioning, We collect a set of high-quality bilingual academic and competitive fine-tuning datasets, including COCO~\cite{lin2014coco}, COCO-CN~\cite{li2019cococn}, Flickr30k~\cite{young2014flickr30k}, Flickr-30k-CNA (consisting of ICM, IQM, ICR, IQR series)~\cite{xie2022flickr30kcna}, AI\_Challenger\_2017~\cite{aichallenger} and TextCaps~\cite{sidorov2020textcaps}. We use a mixture of publicly available data for visual question answering that includes GQA~\cite{hudson2019gqa}, VQAv2~\cite{goyal2017vqav2}, OCR-VQA~\cite{mishra2019ocrvqa}, TextVQA~\cite{singh2019textvqa} and CLEVR~\cite{johnson2017clevr}, where the VQAv2 and GQA datasets are further translated into Chinese through a translation pipeline, named as VQAv2 (T) and GQA (T). Furthermore, for the reference grounding and grounded captioning duality tasks, we construct training instruction-response pairs from Visual Genome~\cite{krishna2017vg}, RefCOCO, RefCOCO+, RefCOCOg~\cite{kazemzadeh2014refercoco} and VCR~\cite{zellers2019vcr}.

\begin{table}[t]
    \centering
    \caption{Details of Ziya-Visual instruction-tuning data in Multi-task Representation Learning stage. The label (T) means that the dataset is initially translated from the original data into Chinese, while the datasets highlighted in red are exclusive to Ziya-Visual-Chat.
    }
\vspace{1em}
\scalebox{1.03}{
    \begin{tabular}{l c c}
         \toprule
         \textbf{Task} & \textbf{\# Samples} & \textbf{Dataset} \\
         \midrule
         Image Captioning     & 8.8M  & \makecell{CC3M, COCO, COCO-CN, Flickr30k, \\
                                            Flickr30k-CNA, AI\_Challenger\_2017, \color{red}TextCaps}\\
                                            \midrule
         Visual Question Answering            & 4.5M  & \makecell{VQAv2, VQAv2 (T), \color{red}GQA, \\ OCR-VQA, GQA (T), TextVQA, CLEVR} \\
         \midrule
         Reference Grounding  & 7.0M  & \makecell{Visual Genome, \color{red}RefCOCO, \\ 
                                            RefCOCO+, RefCOCOg, VCR} \\
                                            \midrule
         Grounded Capation & 7.0M  & \makecell{Visual Genome, \color{red}RefCOCO, \\
                                            RefCOCO+, RefCOCOg, VCR} \\
         \bottomrule
    \end{tabular}
    }
    \label{tab:multitask_data}
\vspace{-1em}  
\end{table}
\begin{table}[t]
    \centering
    \caption{Details of Ziya-Visual instruction-tuning data in Scece-aware Knowledge Learning stage. The label (T) means that the dataset is initially translated from the original data into Chinese, while the datasets highlighted in red are exclusive to Ziya-Visual-Chat.
    }
\vspace{1em}
\scalebox{1}{
    \begin{tabular}{l c c}
         \toprule
         \textbf{Dataset} & Data Description & \textbf{\# Samples} \\
         \midrule
         \multirow{4}{*}{LLaVA} & Captioning & 595k  \\
         & Detail Description & 23k \\
         & Complex Reasoning & 77k \\
         & Conversation & 58k \\
         LLaVA (T) & Total & 753k   \\
         \multirow{7}{*}{M3IT} & Image Captioning & 748k  \\
         & Classification	 & 360k \\
         & Visual Question Answering & 235k \\
         & Knowledgeable Visual QA & 57k \\
         & Reasoning & 121k \\
         & Generation & 174k \\
         & Chinese & 273k \\
         \color{red}M3IT (T) & Total & 1.98M  \\
         \multirow{4}{*}{\color{red}SVIT} & Conversation & 1.6M \\
         & Complex Reasoning	 & 1.6M \\
         & Detail Description & 106k \\
         & Referring QAs	 & 1.0M \\
         \color{red}In-House & Total & 300k \\
         \bottomrule
    \end{tabular}}
    \label{tab:multitask_data}
\vspace{-1em}
\end{table}

\begin{figure}[h]
    \label{tab:instruction-tuning}
  \centering
  \includegraphics[width=\textwidth]{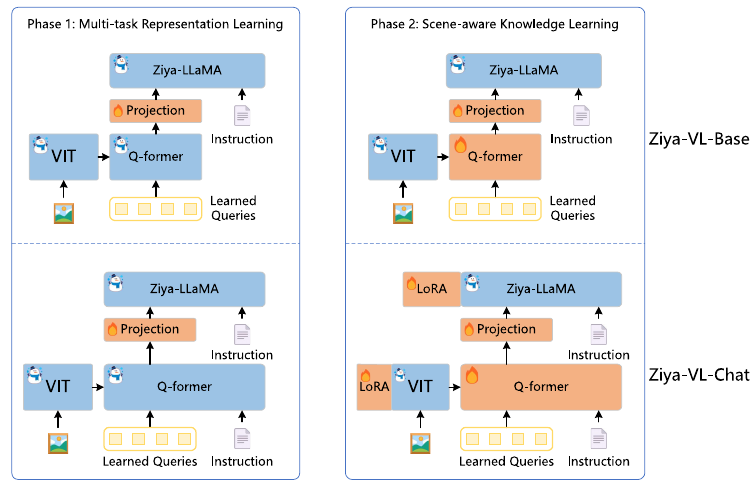}
  \vspace{-1em}
  \caption{The two-stage instruction-tuning pipeline of Ziya-Visual. \textbf{(Top)} The multi-task representation learning and scene-aware knowledge learning for Ziya-Visual-Base. \textbf{(Bottom} The Ziya-Visual-Chat with instruction-aware visual feature extraction and low-rank adaption module.}
\end{figure}

As shown in the left part of Figure~\ref{tab:instruction-tuning}, for Ziya-Visual-Base, we freeze the Q-Former and connect it to the LLM to harvest the LLM’s generative language capability, where the queries transform the image information into understandable text-like representations. A trainable fully-connected (FC) layer is used for linearly projecting the output query tokens into the LLM as instruction prefixes. In Ziya-Visual-Chat, the instruction text is not only given as input to the LLM, but also to the text transformer in Q-Former. The instruction interacts with the queries through self-attention layers of the Q-Former, which influences the queries towards extracting image features that are more informative of the task as described by the instruction~\cite{instructblip}. 

In the second stage of instruction-tuning (Scene-aware Knowledge Learning), we gather a set of high-quality and informative instruction fine-tuning datasets to enhance the instruction following and dialogue capabilities of Ziya-Visual models, encompassing multi-turn conversations, logical reasoning, situational descriptions and detailed depiction in instruction-response scenarios. Therefore, we collect a set of the latest open-source instruction-response datasets, including LLaVA~\cite{liu2023llava}, M3IT~\cite{li2023m3it} and SVIT~\cite{zhao2023svit}, which cover a wide range of multi-modal scenarios in a conversational tone to closely resemble real-world dialogues. Similarly, we also construct a Chinese version of LLaVA (T) and M3IT (T) using the translation pipeline and use the generation pipeline to self-annotate the In-House data.

During training, Ziya-Visual-Base opens up the Q-Former for fine-tuning on top of the existing framework. As for Ziya-Visual-Chat, we consider that the model can learn visual concepts and knowledge depicted in images through visual knowledge learning. Therefore, in addition to opening up the Q-Former and projection layer, we apply low-rank adaptation (LoRA) in the ViT and LLM to adapt $f_V$ and $f_L$ by training multiple low-rank matrices for efficient alignment with human instructions.

%% file: sec/experiment.tex
\section{Experiment And Analysis}

In this section, we evaluate Ziya-Visual series on various traditional vision-language tasks, primarily covering image captioning and general visual question answering (VQA) tasks as presented in Table~\ref{tab:benchmark}. Additionally, we assess the multilingual multi-modal conversation capability of Ziya-Visual in both Chinese and English on the LLaVA benchmark and its Chinese translated version.

\begin{table}[ht]
    \centering
    \caption{Summary of the evaluation benchmarks.}
    \scriptsize
    \vspace{1em}
    \scalebox{1.1}{
    \begin{tabular}{l|l|l|l|l}
        \toprule
         Task & Dataset & Description & Split & Metric  \\
         \midrule
         \multirow{4}{*}{\makecell[l]{Image\\ Caption}} & Nocaps & Captioning of natural images & val & CIDEr($\uparrow$) \\
         & COCO-CN & Captioning of natural images & karpathy-test & CIDEr($\uparrow$) \\
         & Flickr30k-CNA & Captioning of natural images & karpathy-test & CIDEr($\uparrow$) \\
         & Flickr30K & Captioning of natural images & karpathy-test & CIDEr($\uparrow$) \\
         \midrule
         \multirow{5}{*}{\makecell[l]{General\\ VQA}} & VQAv2 & VQA on natural images & test-dev & VQA Score($\uparrow$) \\
         & OKVQA & VQA on natural images requiring outside knowledge & val & VQA Score($\uparrow$) \\
         & GQA & VQA on scene understanding and reasoning & test-balanced & EM($\uparrow$) \\
         & ScienceQA& Multi-choice VQA on a diverse set of science topics & test & Accuracy($\uparrow$) \\
         & VizWiz & VQA on photos taken by people who are blind & test-dev & VQA Score($\uparrow$)\\
         \bottomrule
    \end{tabular}}
    \label{tab:benchmark}
\end{table}

\subsection{Image Captioning and General Visual Question Answering}
Image captioning task requires the model to generate descriptions for a given image. We use Nocaps~\cite{agrawal2019nocaps} and Flickr30k~\cite{young2014flickr30k} as the English evaluation datasets, and report the CIDEr score as metric. As for Chinese version, we choose the common translated subset COCO-CN and Flickr30k-CNA, and extra report the image-text retrieval scores. General VQA task requires model to generate answer for given image-question pair. We utilize five English benchmark for evaluation, including VQAv2~\cite{goyal2017vqav2}, OKVQA~\cite{marino2019okvqa}, GQA~\cite{hudson2019gqa}, ScienceQA (Image Set)~\cite{lu2022scienceqa} and VizWiz VQA~\cite{gurari2018vizwiz}. In general scenarios, we employ open-ended answer generation, while in cases with provided options, we constrain the model's output to the possible choices. We evaluate in a Chinese environment using the GQA (T) and VQAv2 (T) test sets.

\begin{table}[t]
\centering
\caption{Experiment results on English Image Captioning and General VQA.}
\vspace{1em}
\scalebox{0.83}{
\begin{tabular}{@{}l|cc|ccccc@{}}
\toprule
 \multirow{2}{*}{Model} & \multicolumn{2}{c|}{Image Captioning} & \multicolumn{5}{c}{General VQA} \\
   & \begin{tabular}[c]{@{}c@{}}Nocaps\\ (0-shot) \end{tabular} & \begin{tabular}[c]{@{}c@{}}Flickr30K\\ (0-shot) \end{tabular} & VQAv2 & OKVQA & GQA & \begin{tabular}[c]{@{}c@{}}SciQA-Img\\ (0-shot)\end{tabular} & \begin{tabular}[c]{@{}c@{}}VizWiz\\ (0-shot)\end{tabular} \\ \midrule
  Flamingo-9B & - & 61.5 & 51.8 & 44.7 & - & - & 28.8 \\
  Flamingo-80B & - & 67.2 & 56.3 & 50.6 & - & - & 31.6 \\
  Unified-IO-XL & 100.0 & - & 77.9 & 54.0 & - & - & - \\
  Kosmos-1 & - & 67.1 & 51.0 & - & - & - & 29.2 \\
  Kosmos-2 & - & 80.5 & 51.1 & - & - & - & - \\
  BLIP-2 (Vicuna-13B) & 103.9 & 71.6 & 65.0 & 45.9 & 32.3 & 61.0 & 19.6 \\
  InstructBLIP (Vicuna-13B) & \textbf{121.9} & 82.8 & - & - & \textbf{49.5} & 63.1 & 33.4 \\
 Shikra (Vicuna-13B) & - & 73.9 & \textbf{77.4} & 47.2 & - & - & - \\
  \textbf{Ziya-Visual-Base (Ziya-LLaMA-13B)} & 108.2 & 74.7 & 69.2 & 50.3 & 39.8 & 59.8 & 26.6 \\
  \textbf{Ziya-Visual-Chat (Ziya-LLaMA-13B)} & 119.6 & \textbf{83.4} & 75.5 & 47.9 & \textbf{51.1} & \textbf{65.9} & \textbf{34.0} \\  \bottomrule
\end{tabular}
}
\label{tab:en_caption_vqa}
\vspace{-1em}
\end{table}
\begin{table}[t]
\centering
\caption{Experiment results on Chinese Image Captioning and General VQA.}
\vspace{1em}
\scalebox{0.94}{
\begin{tabular}{@{}l|cccccc|ccc@{}}
\toprule
 \multirow{3}{*}{Model} & \multicolumn{6}{c|}{Image Captioning} & \multicolumn{3}{c}{General VQA} \\
   & \multicolumn{3}{c}{COCO-CN} & \multicolumn{3}{c}{Flickr30K} & \multicolumn{2}{c}{GQA (T)} & VQAv2 (T) \\ 
   & CIDEr & IR@1 & TR@1 & CIDEr & IR@1 & TR@1 & ACC & CIDEr & ACC \\ \midrule
  VisualGLM & 87.8 & 85.5 & 94.6 & 86.2 & 82.9 & 89.5 & 32.3 & 93.9 & 45.8 \\
  mPLUG-Owl & 93.7 & \textbf{93.3} & 96.4 & 94.5 & 89.3 & 93.2 & 37.7 & 101.2 & 53.6 \\
  VISCPM-Chat & 108.2 & 91.4 & \textbf{97.7} & \textbf{104.5} & 88.5 & \textbf{93.3} & 41.2 & 113.0 & 58.9 \\
  \textbf{Ziya-Visual-Base} & 95.4 & 89.1 & 94.2 & 91.6 & 86.9 & 90.2 & 36.6 & 104.5 & 54.1 \\
  \textbf{Ziya-Visual-Chat} & \textbf{111.4} & 91.0 & 96.1 & 103.7 & \textbf{91.4} & 92.7 & \textbf{43.4} & \textbf{116.2} & \textbf{60.2} \\  \bottomrule
\end{tabular}
}
\label{tab:zh_caption_vqa}
\end{table}

The overall performance on English image captioning  and general VQA tasks are reported in Table~\ref{tab:en_caption_vqa}. We can observe that Ziya-Visual-Base, adapted for bilingual multi-modal alignment and further instruction-tuning on top of BLIP-2, shows improvement across various English vision-language tasks. Meanwhile, with minimal adjustments using a small amount of Chinese instruction-response pairs, Ziya-Visual-Base demonstrates commendable comprehension and generation abilities in Chinese multi-modal tasks. 
Next, we report the overall performance on Chinese image captioning and general VQA tasks in Table~\ref{tab:zh_caption_vqa}. With the support of more instruction-tuning datasets and comprehensive training strategies, Ziya-Visual-Chat achieves performance on par with English state-of-the-art LVLMs (InstructBLIP~\cite{instructblip} and Shikra~\cite{chen2023shikra}, and outperforms most existing multilingual LVLMs in Chinese scenarios. Specifically, compared to models with outstanding performance in the open-source community, such as mPLUG-Owl~\cite{ye2023mplugowl} and VISCPM-Chat~\cite{hu2023viscpm}, we achieve comparable performance in the image-text retrieval task. Moreover, on the CIDEr metric for the COCO-CN, Flickr30k-CNA and GQA (T) datasets, Ziya-Visual-Chat achieves an average score of 110.4, surpassing mPLUG-Owl's 96.5 and VISCPM-Chat's 108.6. We also observe that on the accuracy metric for GQA (T) and VQAv2 (T), we achieve an average score of 51.8\%, surpassing mPLUG-Owl and VISCPM-Chat by 6.15\% and 1.75\%, respectively. The improvement indicates that Ziya-Visual-Chat can more accurately capture the visual objects within the images via instruction-aware fine-tuning, and low-rank adaption on LLM and ViT modules.

\subsection{LLaVA Benchmark}
LLaVA benchmark consists of 90 instances, each containing an image with symbolic representations, a question and an answer, which comprehensively evaluates the model’s performance in conversation, detailed description and complex reasoning through GPT-4 scoring. We compare Ziya-Visual with existing multi-modal conversation models, which include the English-only models: MiniGPT-4~\cite{zhu2023minigpt4}, InstructBLIP~\cite{instructblip}, and LLaVA~\cite{liu2023llava}, as well as Chinese-English bilingual models: mPLUG-Owl (LLaMA-7B)~\cite{ye2023mplugowl}, VisualGLM (ChatGLM-6B)~\cite{du2022glm} and VISCPM-Chat (CPM-Bee-10B)~\cite{hu2023viscpm}.

As shown in Figure~\ref{tab:llava_result}, Ziya-Visual-Base demonstrates a balanced performance across various metrics, indicating that the model achieves a preliminary multi-modal understanding capability. With the introduction of more bilingual instruction-response pairs, Ziya-Visual-Chat achieves an average GPT-4 score of 84.1 in English and 86.7 in Chinese, surpassing Ziya-Visual-Base by 2.4 and 5.9, respectively. This significant improvement highlights the effectiveness of instruction-aware visual feature extraction and low-rank adaptation strategies in promoting LLM cognition and aligning visual representation, which enables Ziya-Visual-Chat to understand complex instructions and diverse scenarios and provide accurate responses. 

\begin{table}[t]
\centering
\renewcommand\arraystretch{1.25}
\caption{Experimental results on LLaVA benchmark accessed by GPT-4. Con: Conversation, DD: Detailed Description, CR: Complex Reasoning, AVG: the average score of three tasks. The best results are marked in \textbf{bold}.}
\vspace{1em}
\scalebox{0.87}{
\begin{tabular}{cc|c|cccc|cccc}
\toprule
\multicolumn{2}{c|}{\multirow{2}{*}{Model}} & \multirow{2}{*}{\makecell[c]{LLM \\ Backbone}} & \multicolumn{4}{c|}{English}   & \multicolumn{4}{c}{Chinese}  \\ \cline{4-11} 
\multicolumn{2}{c|}{}  &  & \multicolumn{1}{c}{Con} & \multicolumn{1}{c}{DD} & \multicolumn{1}{c|}{CR} & AVG & \multicolumn{1}{c}{Con} & \multicolumn{1}{c}{DD} & \multicolumn{1}{c|}{CR} & AVG \\ \midrule
\multicolumn{1}{c|}{\multirow{3}{*}{\makecell[c]{English \\ Model} }}  & MiniGPT-4 & Vicuna-13B  & \multicolumn{1}{c}{65.0} & \multicolumn{1}{c}{67.3}   & \multicolumn{1}{c|}{76.6}              & 69.7  & \multicolumn{1}{c}{-}   & \multicolumn{1}{c}{-}  & \multicolumn{1}{c|}{-} & - \\ 
\multicolumn{1}{c|}{}  & InstructBLIP  & Vicuna-13B  & \multicolumn{1}{c}{81.9}  & \multicolumn{1}{c}{68.0}   & \multicolumn{1}{c|}{91.2}  & 80.5    & \multicolumn{1}{c}{-}    & \multicolumn{1}{c}{-}  & \multicolumn{1}{c|}{-} & - \\ 
\multicolumn{1}{c|}{} & LLaVA & Vicuna-13B  & \multicolumn{1}{c}{\textbf{89.5}} & \multicolumn{1}{c}{70.4} & \multicolumn{1}{c|}{96.2} & \textbf{85.6}    & \multicolumn{1}{c}{-}  & \multicolumn{1}{c}{-}  & \multicolumn{1}{c|}{-} & - \\ \midrule
\multicolumn{1}{c|}{\multirow{4}{*}{\makecell[c]{\\ \\En-Zh  \\ Bilingual \\ Model}}} & mPLUG-Owl & LLaMA-7B  & \multicolumn{1}{c}{64.6}  & \multicolumn{1}{c}{47.7}  & \multicolumn{1}{c|}{80.1}  & 64.2    & \multicolumn{1}{c}{76.3}         & \multicolumn{1}{c}{61.2}   & \multicolumn{1}{c|}{77.8}    & 72.0      \\ 
\multicolumn{1}{c|}{} & VisualGLM & ChatGLM-6B  & \multicolumn{1}{c}{62.4}  & \multicolumn{1}{c}{63.0}  & \multicolumn{1}{c|}{80.6} & 68.7  & \multicolumn{1}{c}{76.6}  & \multicolumn{1}{c}{87.8} & \multicolumn{1}{c|}{83.6} & 82.7    \\
\multicolumn{1}{c|}{} & \modelname-Chat  & CPM-Bee-10B                             & \multicolumn{1}{c}{81.4}         & \multicolumn{1}{c}{69.2}                 & \multicolumn{1}{c|}{93.1}              & 81.4    & \multicolumn{1}{c}{90.0}         & \multicolumn{1}{c}{\textbf{87.4}}                 & \multicolumn{1}{c|}{\textbf{95.0}}              & \textbf{90.9} \\  
\cline{2-11}
\multicolumn{1}{c|}{}  & Ziya-Visual-Base & Ziya-LLaMA-13B & \multicolumn{1}{c}{82.7}  & \multicolumn{1}{c}{69.9}  & \multicolumn{1}{c|}{92.1}  & 81.7    & \multicolumn{1}{c}{85.0}  & \multicolumn{1}{c}{74.7}     & \multicolumn{1}{c|}{82.4} & 80.8  \\ 
\multicolumn{1}{c|}{}  & Ziya-Visual-Chat & Ziya-LLaMA-13B & \multicolumn{1}{c}{84.2}  & \multicolumn{1}{c}{\textbf{71.4}}  & \multicolumn{1}{c|}{\textbf{96.6}}  & 84.1    & \multicolumn{1}{c}{\textbf{90.9}}  & \multicolumn{1}{c}{81.3}     & \multicolumn{1}{c|}{88.0} & 86.7 
   \\ \bottomrule 
\end{tabular}}
\label{tab:llava_result}
\vspace{-1em}
\end{table}

\subsection{Case Study}

\begin{figure}[h]
    \label{tab:case_study_2}
  \centering
  \includegraphics[width=\textwidth]{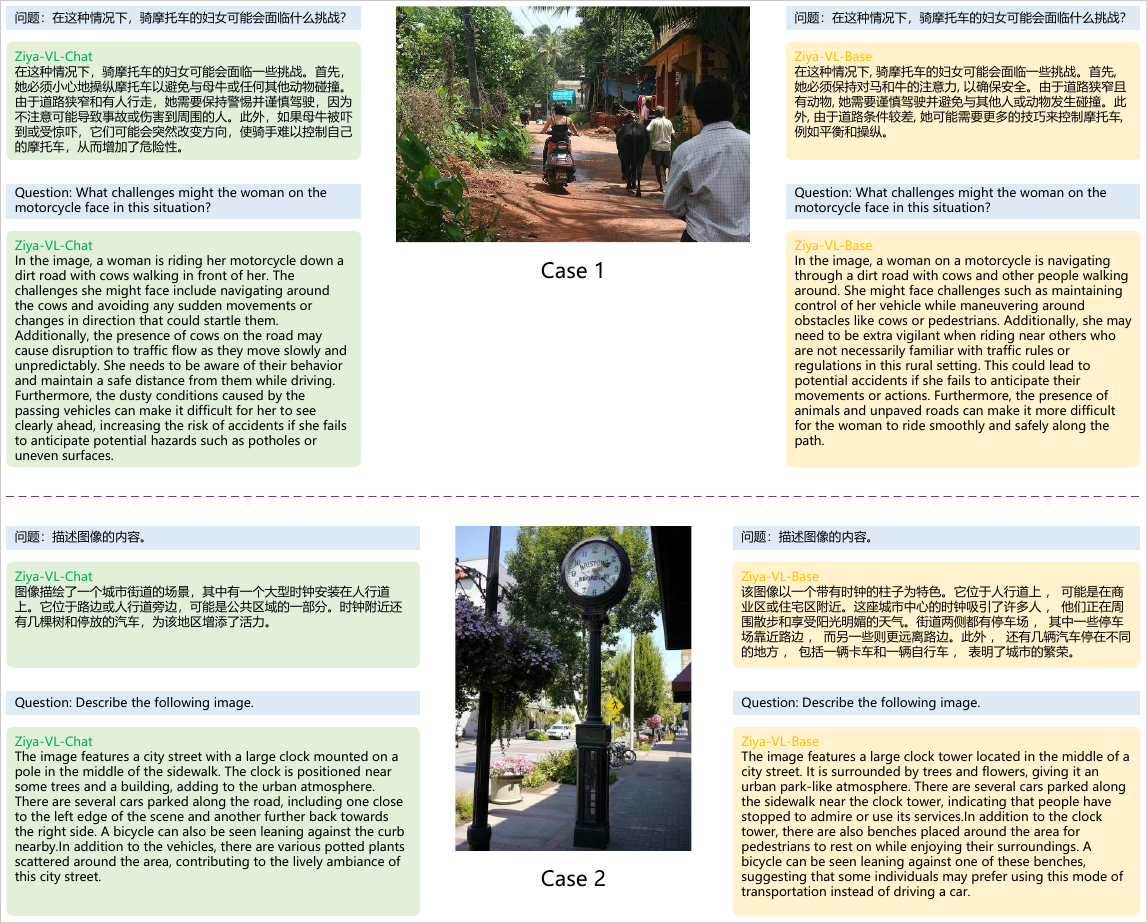}
  \vspace{-1em}
  \caption{Case Study of Ziya-Visual-Base and Ziya-Visual-Chat on bilingual multi-modal generation.}
\end{figure}

To intuitively observe the multi-modal understanding and generation capabilities of the Ziya-Visual series, in Figure~\ref{tab:case_study_2}, we present dialogues involving Ziya-Visual-Base and Ziya-Visual-Chat in a selection of qualitative examples. 

From the prediction results of case 1 complex reasoning and case 2 detail description, we can observe that Ziya-Visual-Base demonstrate initial multi-modal understanding and generation capabilities. It enables the model to comprehensively perceive visual objects and global semantic in images, further providing corresponding responses to instructions. However, Ziya-Visual-Base still exhibits significant deficiencies in visual hallucination and expression fluency. Specifically, in the responses for case 1 and case 2, it erroneously mentioned\begin{CJK*}{UTF8}{gbsn} '马(horse)' and '卡车(truck)' 
 \end{CJK*} respectively, which are not visual objects present in the images. Furthermore, even though the model can generate answers with long context, it still misses many elements in the images.

Excitingly, Ziya-Visual-Chat significantly addresses the aforementioned shortcomings, which can accurately capture image details and provide a well-structured analysis and reasoning chain based on instructions. We can observe that in case 1, Ziya-Visual-Chat can provide a well-founded analysis of the 'What challenges might the woman on the motorcycle face in this situation?' from various aspects. In case 2, although the model's answer is concise, it accurately covers visual objects such as \begin{CJK*}{UTF8}{gbsn} '城市街道(city street)', '大型时钟(large clock)', '树(tree)' and '汽车(car)'. \end{CJK*}

